\documentclass[letterpaper, 10 pt, conference]{ieeeconf}  

\IEEEoverridecommandlockouts                              

\usepackage{amssymb}  
\usepackage{cite}

\title{\LARGE \bf
\textbf{Kilohertz-Safe:} A Scalable  Framework for Constrained \\ Dexterous Retargeting}

\author{Yinxiao Tian$^{*}$, Ziyi Yang$^{*}$ Zinan Zhao and Zhen Kan
\thanks{$^{*}$Equal contribution.}%
\thanks{Y. Tian, Z. Yang, Z. Zhao and Z. Kan are with the Department of Automation at the University of Science and Technology of China, Hefei, Anhui, China, 230026.}%
}

\usepackage{graphicx}
\usepackage{amsmath}
\usepackage{threeparttable}
\usepackage{booktabs}
\usepackage{multirow}
\usepackage{siunitx}
\usepackage{xcolor}
\usepackage{float} 
\newtheorem{remark}{Remark}
\usepackage{hyperref}
\begin{document}

\maketitle
\thispagestyle{empty}
\pagestyle{empty}

\begin{abstract}

Dexterous hand teleoperation requires motion retargeting methods that simultaneously achieve high-frequency real-time performance and enforcement of heterogeneous kinematic and safety constraints. Existing nonlinear optimization–based approaches often incur prohibitive computational cost, limiting their applicability to kilohertz-level control, while learning-based methods typically lack formal safety guarantees. This paper proposes a scalable motion retargeting framework that reformulates the nonlinear retargeting problem into a convex quadratic program in joint differential space. Heterogeneous constraints, including kinematic limits and collision avoidance, are incorporated through systematic linearization, resulting in improved computational efficiency and numerical stability. Control barrier functions are further integrated to provide formal safety guarantees during the retargeting process. The proposed framework is validated through simulations and hardware experiments on the Wuji Hand platform, outperforming state-of-the-art methods such as Dex-Retargeting and GeoRT. The framework achieves high-frequency operation with an average latency of 9.05 ms, while over 95\% of retargeted frames satisfy the safety criteria, effectively mitigating self-collision and penetration during complex manipulation tasks.
\end{abstract}

\section{INTRODUCTION}
Dexterous teleoperation requires mapping high-dimensional human hand motions to robotic hands under strict real-time and safety constraints. In practical systems, control commands must be generated at millisecond-level latency to maintain responsiveness and closed-loop stability. Meanwhile, dexterous robotic hands are mechanically compact, with millimeter-scale clearances between fingers and links, making them highly sensitive to numerical errors, control delays, and constraint violations. Even small deviations can result in self-collisions, actuator saturation, or hardware damage. Consequently, motion retargeting for dexterous manipulation must simultaneously support high-frequency execution and strict enforcement of heterogeneous constraints.

Existing dexterous hand retargeting methods generally follow two paradigms. Optimization-based approaches, such as DexPilot \cite{9197124} and AnyTeleop \cite{qin2023anyteleop}, formulate retargeting as constrained optimization problems that explicitly encode kinematic consistency, joint limits, and motion smoothness. These methods are interpretable and adaptable across different robotic hands, but typically rely on iterative nonlinear solvers whose computational cost scales poorly with increasing degrees of freedom and constraint complexity, limiting their ability to achieve stable kilohertz-level control.

Learning-based approaches address computational limitations by shifting complexity to offline training. Recent methods such as GeoRT \cite{yin2025geort} demonstrate that direct human-to-robot mappings can achieve retargeting at 1 kHz with low inference latency. However, such approaches are often tightly coupled to specific robot embodiments and training datasets. Safety constraints are usually enforced implicitly through loss functions or heuristics, providing no formal guarantees. As a result, performance can degrade in out-of-distribution configurations or contact-rich scenarios, which restricts their use in safety-critical manipulation tasks.

Formal safety and feasibility guarantees have been widely studied in control and manipulation. In particular, Control Barrier Functions (CBFs) provide a principled mechanism for enforcing state and input constraints with theoretical guarantees in real time \cite{11267490,ames2019control,xu2020robust}. However, integrating such guarantees into high-frequency dexterous hand retargeting remains challenging. Most existing retargeting pipelines rely on heuristic collision penalties or offline nonlinear optimization \cite{rakita2021collisionik,gaertner2021collisionfree,11246607,cheng2024open,wang2024dexcap,naughton2024respilot}, and lack a unified, computationally efficient framework for incorporating kinematic, geometric, and safety constraints at kilohertz-level control rates.

To address this challenge, we propose Kilohertz-Safe, a scalable dexterous teleoperation retargeting framework designed from a system architecture perspective. Instead of formulating retargeting as a standalone nonlinear optimization or a purely learned mapping, Kilohertz-Safe introduces a linearized control interface in joint differential space, which enables heterogeneous constraints to be consistently expressed and enforced under high-frequency control. Within this unified formulation, safety is treated as a first-class constraint rather than an auxiliary objective, and collision avoidance is enforced with formal guarantees. The resulting framework achieves computational efficiency suitable for stable millisecond-level control in high-dimensional dexterous hand systems.

\begin{itemize}
    \item We present Kilohertz-Safe, a scalable dexterous teleoperation retargeting framework that reconciles high-frequency real-time execution with heterogeneous constraint satisfaction, enabling stable millisecond-level control for high-DoF robotic hands.
    \item We formulate safety specifications as explicit constraints within a unified joint differential-space interface, providing formal guarantees of collision avoidance while preserving computational efficiency.
     \item We validate the proposed framework through real-world experiments on a dexterous hand platform, demonstrating its real-time performance and practical effectiveness under safety-critical manipulation scenarios.
\end{itemize}

\begin{figure*}[!htb]
  \centering
  \includegraphics[width=0.85\textwidth]{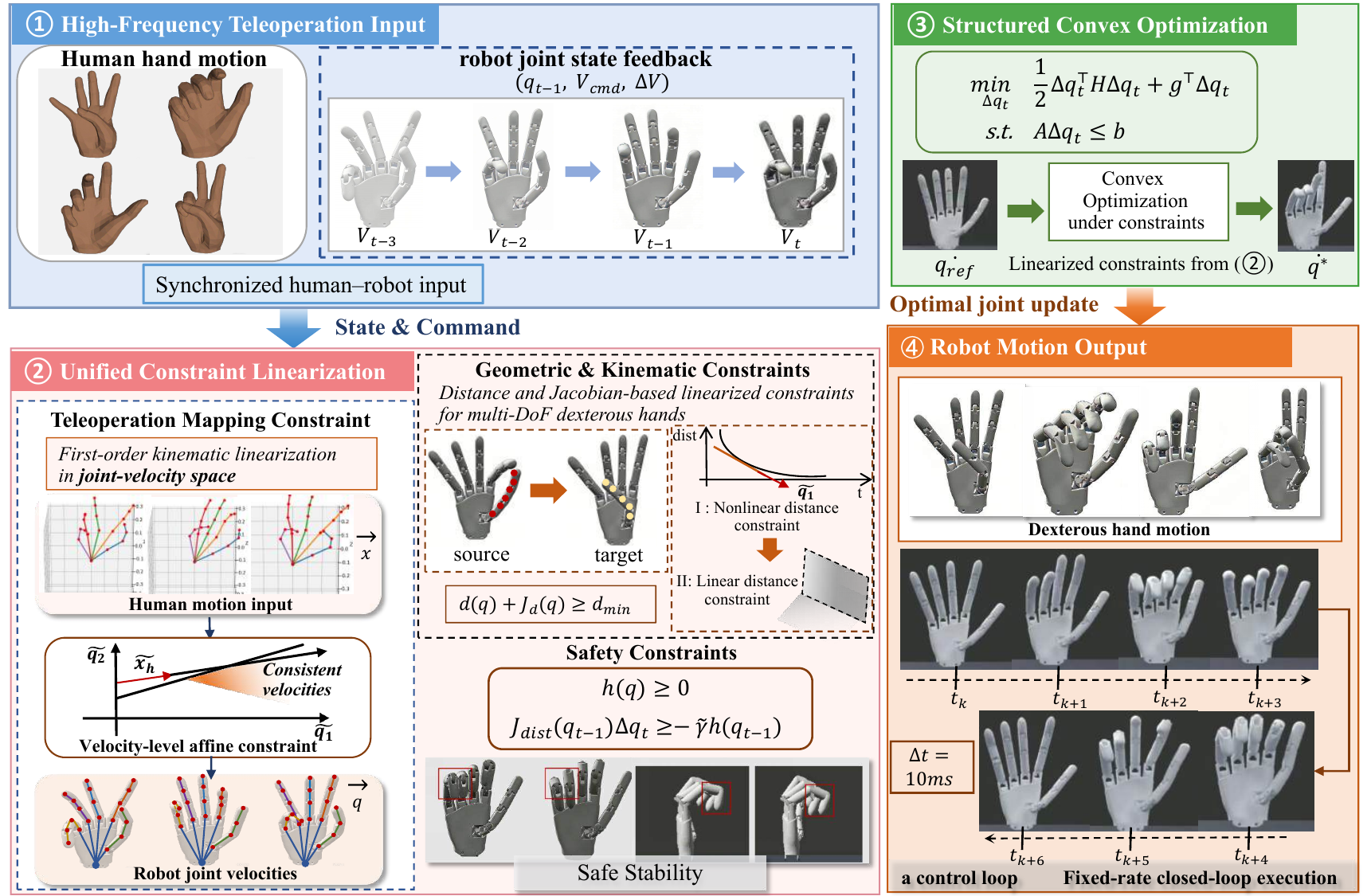}
  \caption{\textbf{System pipeline of the proposed high-frequency retargeting framework.}
High-frequency human inputs and robot state feedback are unified through velocity-level constraint linearization, forming a structured convex optimization problem that enables stable real-time retargeting under geometric, kinematic, and safety constraints.}
  \label{framework}
\end{figure*}

\section{Preliminaries}
\subsection{Dexterous Hand Retargeting}
Dexterous hand retargeting aims to map human hand motion to robot joint
configurations such that the resulting robot hand pose resembles the human
hand pose while respecting the kinematic limits of the robot.
Due to the high dimensionality of the hand and the morphological differences between human and robot embodiments, this mapping is commonly formulated as an optimization problem.

Specifically, geometric features such as fingertip positions or joint keypoints are often used to characterize the hand pose. Let $\mathbf{v}_i^t \in \mathbb{R}^3$, $i = 1, \ldots, N$, denote the 3D position of the $i$-th human hand keypoint at time step $t$, expressed in a common observation frame, where $N$ is the total number of selected hand keypoints.
The retargeting problem aims to compute a robot joint configuration
$\mathbf{q}_t \in \mathbb{R}^{n_q}$ whose forward kinematics aligns the
corresponding robot keypoints with the observed human hand features. A representative optimization problem commonly adopted in prior works  \cite{qin2023anyteleop,9197124,qin2021dexmv,pyroki2025modular} is formulated as
\begin{equation}
\begin{aligned}
\min_{\mathbf{q}_t} \quad &
\sum_{i=1}^{N} \alpha \left\lVert \mathbf{v}_i^{t}
- \mathbf{f}_i(\mathbf{q}_t) \right\rVert^{2}
+ \beta \left\lVert \mathbf{q}_t - \mathbf{q}_{t-1} \right\rVert^{2} \\
\text{s.t.} \quad &
\mathbf{q}_l \le \mathbf{q}_t \le \mathbf{q}_u ,
\end{aligned}
\label{ori_tar}
\end{equation}
where $\mathbf{f}_i(\cdot): \mathbb{R}^{n_q} \rightarrow \mathbb{R}^3$ denotes the forward kinematics mapping from the robot joint configuration to the position of the $i$-th robot keypoint. The weighting coefficients
$\alpha,\beta\in\mathbb{R}^{+}$ balance task-space tracking accuracy and temporal
smoothness, while the joint limits $\mathbf{q}_l$ and $\mathbf{q}_u$ enforce kinematic feasibility.

This formulation captures the core objective of dexterous hand retargeting:
aligning human and robot hand task-space features, such as fingertips, through a kinematically feasible robot joint configuration.

\subsection{Control Barrier Functions}
Control Barrier Functions (CBFs) \cite{ames2017control} provide a principled way to enforce safety constraints in continuous-time control systems.
Given a system state $\mathbf{x} \in \mathbb{R}^n$, consider a safety set 
\begin{equation}
\mathcal{C} = \left\{ \mathbf{x} \mid h(\mathbf{x}) \ge 0 \right\},
\end{equation}
where $h(\mathbf{x})$ is a continuously differentiable function that encodes safety constraints.

To ensure forward invariance of the safety set, a standard CBF condition can be imposed on the system dynamics.
For first-order systems, this condition takes the form \cite{xu2020robust,ames2017control}
\begin{equation}
\dot{h}(\mathbf{x}) + \alpha h(\mathbf{x}) \ge 0,
\end{equation}
where $\alpha > 0$ is a user-defined constant.
This condition guarantees that the system state remains within the safe set
$\mathcal{C}$ over time.

In optimization-based control, the CBF condition can be expressed as an inequality constraint on the control input \cite{ames2017control}. When applied to velocity-level or differential control of robotic systems, the above condition can be written as a linear constraint with respect to the joint velocity or
increment\cite{7526486},
\begin{equation}
\nabla h(\mathbf{x})^\top \dot{\mathbf{q}} \ge -\alpha h(\mathbf{x}),
\end{equation}
which enables real-time enforcement of safety constraints during control.
This makes CBFs a suitable mechanism for incorporating formal safety guarantees into dexterous hand retargeting systems.

\section{Method}
This section presents a retargeting framework for dexterous hand teleoperation that enables stable and predictable real-time execution at kilohertz control rates. Within this framework, teleoperation mappings, geometric and kinematic constraints, as well as safety-related constraints, although inherently nonlinear in joint space, are locally linearized around the current operating point and expressed as linear equality and inequality constraints on joint velocities. This unified formulation recasts dexterous-hand retargeting as a structured convex optimization problem with bounded and predictable computational complexity. As a result, the proposed framework is well suited for high-frequency control loops and remains scalable across dexterous hands with different kinematic structures and degrees of freedom. The remainder of this section details the individual components of the framework and their practical implementation.

\subsection{Quadratic Programming Formulation of Hand Pose Retargeting}
Following AnyTeleop \cite{qin2023anyteleop} and most existing retargeting approaches \cite{sharma2018dexterous,9197124,wen2025bytedexter,park2025learning,zhao2025dexmachina}, as formulated in (1), we model the hand pose retargeting as a nonlinear optimization problem defined in joint space. The objective is to minimize the discrepancy between human and robot hand keypoints, while simultaneously enforcing joint position limits and temporal smoothness constraints. Specifically, given the joint configuration increment at time $t$
\[ \Delta \mathbf{q}_t= \mathbf{q}_t - \mathbf{q}_{t-1} \in \mathbb{R}^n, \] applying a first-order Taylor expansion to the forward kinematics at $\mathbf{q}_{t-1}$ yields
\begin{equation}
f_i(\mathbf{q}_t) \approx f_i(\mathbf{q}_{t-1}) + J_i(\mathbf{q}_{t-1}) \, \Delta \mathbf{q}_t ,
\label{zhankai}
\end{equation}
where $J_i$ is the Jacobian of the $i$-th keypoint with respect to the joint variables.
Substituting \eqref{zhankai} into \eqref{ori_tar} leads to a
quadratic objective in $\Delta \mathbf{q}_t$ as
\begin{equation}
   \min_{\Delta \mathbf{q}_t}  \| \mathbf{J} \Delta \mathbf{q}_t - \Delta \mathbf{v} \|_2^2
+ \beta \| \Delta \mathbf{q}_t \|_2^2
\end{equation}
where
\[
J =
\begin{bmatrix}
J_1 \\
\vdots \\
J_N
\end{bmatrix},
\qquad
\Delta \mathbf{v} =
\begin{bmatrix}
\Delta \mathbf{v}_1^{t} \\
\vdots \\
\Delta \mathbf{v}_N^{t}
\end{bmatrix},
\Delta \mathbf{v}_i^{t} = \mathbf{v}_i^{t} - f_i(\mathbf{q}_{t-1})
\]
Meanwhile, the original joint position limits
$\mathbf{q}_l \le \mathbf{q}_t \le \mathbf{q}_u$
can be rewritten as linear inequality constraints on the joint differential as
\[
\mathbf{q}_l - \mathbf{q}_{t-1} \le \Delta \mathbf{q}_t \le \mathbf{q}_u - \mathbf{q}_{t-1}.
\]
As a result, the original nonlinear retargeting problem in literature (e.g., AnyTeleop) is transformed into a convex quadratic program with respect to the joint differential. The resulting standard QP form is
\begin{equation}
\begin{aligned}
\min_{\Delta \mathbf{q}_t} \quad & \frac{1}{2} \Delta \mathbf{q}_t^\top H \Delta \mathbf{q}_t + \mathbf{g}^\top \Delta \mathbf{q}_t \\
\text{s.t.} \quad & \mathbf{A} \Delta \mathbf{q}_t \le \mathbf{b} ,
\label{QP}
\end{aligned}
\end{equation}
where 
\[
\mathbf{H} = 2 \left( \alpha \mathbf{J}^\top \mathbf{J} + \beta \mathbf{I} \right),
\mathbf{g} = -2 \alpha \mathbf{J}^\top \Delta \mathbf{v} \]
\[
\mathbf{A} =
\begin{bmatrix}
\mathbf{I} \\
- \mathbf{I}
\end{bmatrix},
\qquad
\mathbf{b} =
\begin{bmatrix}
\mathbf{q}_u - \mathbf{q}_{t-1} \\
\mathbf{q}_{t-1} - \mathbf{q}_l
\end{bmatrix}.
\]
By construction (\ref{QP}) has a positive semi-definite Hessian, guaranteeing convexity and enabling reliable real-time solutions.
This QP formulation also serves as a unified backbone for incorporating additional kinematic, geometric, and safety-related constraints, as discussed in the following section.

\begin{remark}
With the availability of virtual reality interfaces, vision-based tracking, and data gloves, human hand motion can now be acquired reliably at high frequency. This enables dexterous-hand teleoperation to be executed within a high-rate control loop, where the variation in joint configurations between consecutive time steps is inherently small.
\end{remark}

\subsection{Control Barrier Function–Based Safety Guarantee}
While the quadratic program in \eqref{QP} enables task objectives and physical constraints to be handled in a unified manner under high-frequency control, it does not by itself guarantee satisfaction of safety constraints throughout execution. Existing nonlinear optimization based retargeting approaches typically encode collision avoidance by adding distance or collision related penalties to the objective function. Such treatments effectively impose soft constraints, whose safety performance depends critically on heuristic weight tuning and therefore can only mitigate collision risks in a probabilistic or empirical sense, rather than providing formal guarantees.

To enforce safety, we incorporate CBFs into the retargeting formulation, yielding provable forward invariance of the safe set during execution. Specifically, we construct a capsule-based collision model in which robot links and environmental obstacles are approximated by sets of geometric primitives. For any potentially colliding pair of bodies $A$ and $B$, let $\mathbf{p}_A(\mathbf{q}), \mathbf{p}_B(\mathbf{q}) \in \mathbb{R}^3$ denote the closest (witness) points on their respective skeletal segments. The corresponding safety function is defined as
\begin{equation}
h(\mathbf{q}) = \|\mathbf{p}_A(\mathbf{q}) - \mathbf{p}_B(\mathbf{q})\|_2 - (r_A + r_B),
\end{equation}
where $r_A$ and $r_B$ are the radii of capsules $A$ and $B$, respectively. A collision occurs when $h(\mathbf{q}) < 0$, and the safe configuration set is given by $\mathcal{S}=\{\mathbf{q}\in \mathbb{R}^n \mid h(\mathbf{q})\ge 0\}$. 

To guarantee forward invariance of $\mathcal{S}$, we impose the condition $\dot{h}(\mathbf{q}) \ge -\gamma h(\mathbf{q}),$ where $\gamma > 0$ controls the rate at which the system is driven away from the safety boundary. Enforcing this condition requires evaluating the time derivative of $h(\mathbf{q})$, which can be obtained via differential kinematics and Danskin’s Theorem. Let 
\[\hat{\mathbf{n}} = \frac{\mathbf{p}_A - \mathbf{p}_B}{\|\mathbf{p}_A - \mathbf{p}_B\|_2}\] 
denote the unit normal vector pointing from body $B$ to body $A$. The linear velocities of the witness points in the world frame are 
\[\dot{\mathbf{p}}_A = \mathbf{J}_{v,A}(\mathbf{q}; \mathbf{p}_A)\dot{\mathbf{q}},\]  \[\dot{\mathbf{p}}_B = \mathbf{J}_{v,B}(\mathbf{q}; \mathbf{p}_B)\dot{\mathbf{q}},\]
where $\mathbf{J}_{v,A}, \mathbf{J}_{v,B} \in \mathbb{R}^{3 \times n}$ are the corresponding point Jacobians. Applying the chain rule yields
\[
\dot{h}(\mathbf{q})=\frac{\partial h}{\partial\mathbf{q}}\dot{\mathbf{q}}=\hat{\mathbf{n}}^{\top}(\dot{\mathbf{p}}_{A}-\dot{\mathbf{p}}_{B})=\mathbf{J}_{\mathrm{dist}}(\mathbf{q})\dot{\mathbf{q}},
\]
where the distance Jacobian is defined as
\[
\mathbf{J}_{\mathrm{dist}}(\mathbf{q}) \triangleq \hat{\mathbf{n}}^\top (\mathbf{J}_{v,A}(\mathbf{q}; \mathbf{p}_A) - \mathbf{J}_{v,B}(\mathbf{q}; \mathbf{p}_B)),
\]
which establishes a linear mapping between joint velocities and variations of the inter-body distance.

Under high-frequency teleoperation, a first-order discretization $\dot{\mathbf{q}}\approx \Delta\mathbf{q}_t/\Delta t$ leads to the linear inequality
\[\mathbf{J}_{\mathrm{dist}}(\mathbf{q}_{t-1})\,\Delta\mathbf{q}_t\ge -\tilde{\gamma}\,h(\mathbf{q}_{t-1}),\] where $\tilde{\gamma}=\gamma\triangle t$.  
This constraint is affine in the decision variable $\triangle\mathbf{q}_{t}$ and can be incorporated directly into the quadratic program alongside joint limit constraints. Specifically, the stacked inequality matrices are constructed as $\mathbf{A}=[\mathbf{A}_{\mathrm{jl}};-\mathbf{J}_{\mathrm{dist}}]$ and $\mathbf{b}=[\mathbf{b}_{\mathrm{jl}};\tilde{\gamma}\,\mathbf{h}(\mathbf{q}_{t-1})]$. 
Because both the objective function and all constraints remain convex, the resulting optimization problem is a standard convex QP whose feasible set is the intersection of convex sets. Importantly, this formulation provides formal geometric safety guarantees for dexterous hand teleoperation while preserving real-time solvability. In contrast, directly embedding CBF constraints into a nonlinear retargeting formulation would introduce high-dimensional nonlinear inequalities, leading to prohibitive computational costs and rendering high-frequency execution impractical.

\section{Experiment}

\subsection{Experimental Setup}
\label{Sec IV.A}
We evaluate the proposed method in both simulation and real-world experiments on the Wuji dexterous hand platform. The simulation is conducted in the SAPIEN \cite{xiang2020sapiensimulatedpartbasedinteractive} environment using the official URDF model of the Wuji hand. Human hand keypoints are captured using a monocular laptop camera and processed by the proposed retargeting pipeline to output joint angles, which are executed directly in simulation. This setup enables controlled evaluation of motion fidelity, safety behavior, and computational performance.

For hardware validation, the Wuji dexterous hand is mounted on a fixed platform. Human hand motion is captured using an Intel RealSense camera, and hand keypoints together with wrist poses are estimated in real time via MediaPipe \cite{lugaresi2019mediapipe}. Both hand keypoints and wrist poses are estimated in real time and retargeted to the robot hand, demonstrating the applicability of the proposed framework under practical sensing and actuation conditions.

We compare our approach against two representative retargeting methods: Dex-Retargeting\cite{qin2023anyteleop} and GeoRT\cite{yin2025geort}, under identical task setups and environmental conditions. Performance is assessed using the following frame-wise quantitative metrics.

\begin{itemize}
\item[\hspace{1em} 1)] \textbf{Computation Latency:} We evaluate computational efficiency by measuring the per-step latency of each pipeline, defined as the wall-clock time required to map a single human hand input frame to the corresponding robot joint targets. The reported latency includes hand pose processing, retargeting computation, and constraint-aware optimization, while excluding rendering, visualization, and communication overhead.
    
    Latency is recorded at each control step using system timestamps. We report the mean, standard deviation, and 99th percentile across all trials. In addition, we report the percentage of control steps that meet the target control frequency, which reflects the real-time feasibility of each method under a fixed control period.
    \item[\hspace{1em} 2)] {\textbf{Motion Preservation:}} Motion preservation measures the directional consistency between the human and robot hands. Specifically, we uniformly sample $N$ anchor points $\{x_1, x_2, \dots, x_N\}$ on the hand surface. At each anchor point $x_i$, we compute the directional dissimilarity $\epsilon_i = 1-(d_i^H)^\top d_i^R$, where $d_i^H$ and $d_i^R$ denote the corresponding unit directional vectors of the human and robot hands, respectively. The value of $\epsilon_i$ lies in the range $[0, 2]$, where 0 indicates perfect alignment and 2 indicates opposite directions. The overall motion preservation at time step $t$ is defined as
    \[
    E_t = \frac{1}{N} \sum_{i=1}^{N} w_i \cdot \epsilon_i,
    \]
    where $w_i$ is a predefined weight associated with anchor point $x_i$. 
Values closer to zero indicate stronger directional agreement and better motion preservation.

    \item[\hspace{1em} 3)] \textbf{Collision Safety Score:} To quantify self-collision avoidance among the fingers of the dexterous hand, we evaluate the minimum geometric clearance between non-adjacent finger links. We refer to this quantity as the inter-finger distance, which captures the closest approach between any pair of fingers during execution. Formally, the minimum self-collision distance over a trial is defined as
    \[
    D_{\mathrm{self}} = \min_{t \in \mathcal{T}} \, \min_{(i,j)\in\mathcal{P}} 
    \mathrm{dist}\!\left( \mathbf{q}_i(t), \mathbf{q}_j(t) \right),
    \]
    where $\mathbf{q}_i(t)$ and $\mathbf{q}_j(t)$ denote the configurations of the $i$-th and $j$-th links at time $t$, $\mathcal{T}$ denotes the evaluation horizon of the trial, and $\mathcal{P}$ denotes the set of non-adjacent link pairs considered. 
    The function $\mathrm{dist}(\cdot)$ computes the minimum Euclidean distance between the corresponding link geometries.
    
    For normalized comparison across trials, we define a safety score
    \[
    S_{\mathrm{safe}} = \mathrm{clip}\!\left( \frac{D_{\mathrm{self}}}{D_{\mathrm{safe}}}, \, 0, \, 1 \right),
    \]
    where $D_{\mathrm{safe}}$ is a predefined clearance threshold. This metric serves as a relative indicator of geometric self-collision safety rather than a complete physical safety model.

\end{itemize}

\subsection{Simulation Evaluation}
We first evaluate different retargeting methods in the simulation environment. The evaluation covers motion preservation, collision safety, and real-time performance, and further includes an ablation study to isolate the effect of explicit safety constraints. For quantitative assessment, we report computation latency, as well as the Motion Preservation and Collision Safety Score introduced in Sec \ref{Sec IV.A}.
\begin{table}[t]
  \centering
  \footnotesize 
  \setlength{\tabcolsep}{2.7pt}
  
  \renewcommand{\arraystretch}{1.15}
  \caption{Latency Comparison}
  \begin{tabular}{l
                  S[table-format=2.2]
                  S[table-format=2.2]
                  S[table-format=2.2]
                  S[table-format=2.2]}
    \toprule
    Method
    & {Mean (ms)$\downarrow$}
    & {Std (ms)$\downarrow$}
    & {99\%ile (ms)$\downarrow$}
    & {RT@100Hz (\%)$\uparrow$}\\
    \midrule
    \textbf{Ours}
      & \textbf{9.05}
      & \textbf{2.29}
      & \textbf{13.42}
      & \textbf{85.82}\\
    Dex-Retargeting
      & 15.59
      & 12.50
      & 32.82
      & 34.41\\
    GeoRT 
      & 34.49
      & 4.28
      & 49.90
      & 0.19 \\
    \bottomrule
  \end{tabular}
  
  \begin{tablenotes}
  \item[*]Runtime latency comparison of teleoperation pipelines on the simulated Wuji hand platform. Latency is measured as the per-step wall-clock computation time required to map a single human hand input frame to robot joint targets, excluding rendering and visualization. RT@100Hz denotes the percentage of control steps whose computation latency is below the 10\,ms control period.
  \end{tablenotes}
  \label{tab:latency}
\end{table}
\subsubsection{Computation Latency}
Table \ref{tab:latency} summarizes the real-time performance comparison of three retargeting methods on a simulated Wuji Hand platform. Owing to its unified differential-space formulation, our approach achieves lower average computation latency, reduced latency variability, and a higher proportion of control steps satisfying the target control frequency. These results demonstrate that the proposed system architecture enables stable millisecond-level retargeting under heterogeneous constraints. In contrast, GeoRT exhibits higher computation latency, which is associated with its reliance on geometric mapping operations and per-frame heuristic corrections. Dex-Retargeting shows larger latency fluctuations, as its optimization process exhibits input-dependent computational complexity, leading to less consistent real-time behavior.

\begin{figure*}[!htb]
  \centering
  \includegraphics[width=\textwidth]{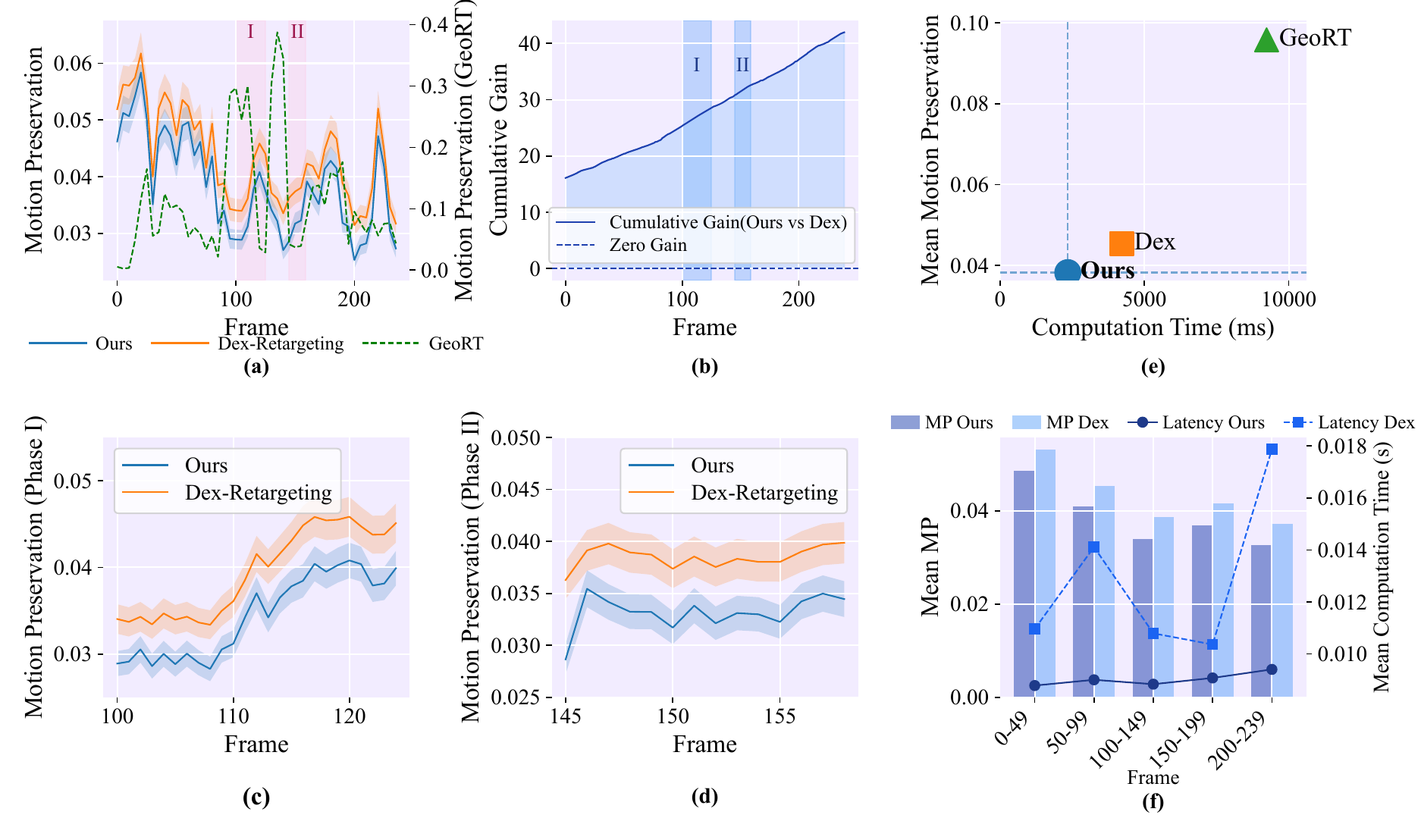}
  \caption{\textbf{Motion preservation comparison across retargeting pipelines.}
(a) Global motion error over the entire interaction sequence. Ours and Dex-Retargeting are plotted against the left vertical axis, whereas GeoRT is shown using the right vertical axis due to its distinct numerical scale. Dual axes are adopted solely for visualization clarity.
(b) Cumulative motion preservation advantage of our method compared with Dex-Retargeting.
(c–d) Zoomed-in views of critical interaction stages, highlighting fine-grained finger posture preservation.
(e) Efficiency–fidelity trade-off: mean motion preservation, averaged over all frames in the interaction sequence, versus total computation time accumulated over the same sequence for Ours, Dex-Retargeting, and GeoRT.
(f) Segment-wise comparison (50-frame bins) of mean motion preservation and mean computation time: bars for motion preservation, markers for latency.}
  \label{fig2}
\end{figure*}
\subsubsection{Motion Preservation}
Regarding motion preservation, the proposed retargeting formulation maintains closer alignment with the human hand motions throughout the interaction sequence, as illustrated in Fig. \ref{fig2}(a). Dex-Retargeting demonstrates noticeable deviations in certain interaction phases, while GeoRT shows reduced motion fidelity due to geometric approximations and discontinuous per-frame corrections. We define the cumulative motion preservation error for method $k$ as
\[
E_c^{(k)}(T) = \sum_{t=1}^{T} E^{(k)}(t),
\]
where $E^{(k)}(t)$ denotes the instantaneous motion deviation from the reference at time step $t$. $T$ denotes the fixed evaluation horizon shared by all methods. The \textit{Cumulative Motion Preservation Gain} is defined in a normalized form as
\[
G_{\mathrm{rel}}(T) =
\frac{
E_c^{(\mathrm{baseline})}(T) - E_c^{(\mathrm{ours})}(T)
}{
E_c^{(\mathrm{baseline})}(T)
}.
\]
where \textit{baseline} denotes the compared retargeting method. Positive values indicate improved motion preservation.

For visualization clarity, Fig.~\ref{fig2}(b) reports the cumulative gain between our method and Dex-Retargeting. As time progresses, the cumulative gain increases steadily, indicating that our method consistently accumulates less motion deviation over extended horizons. 
This demonstrates improved long-term motion fidelity during retargeting. Zoomed-in views of Phase I, where finger crossing occurs, and Phase II, where the fingers close to form a stable grasp (consistent with the bottom-left motion in Fig.~3), as shown in Fig.~\ref{fig2}(c–d), further illustrate these differences. Fig.~\ref{fig2}(e) shows the relationship between total computation time and mean motion preservation (MP) for the three methods, where MP is averaged over all frames in the interaction sequence. Across the interaction sequence, our method achieves the lowest MP values while maintaining minimal computation time. Fig.~\ref{fig2}(f) provides a segment-wise analysis (50-frame bins), revealing that our method not only preserves high-frequency responsiveness but also improves directional consistency within each segment.

\begin{figure}[H]
  \centering
  \hspace*{-0.5cm} 
  \includegraphics[width=0.53\textwidth]{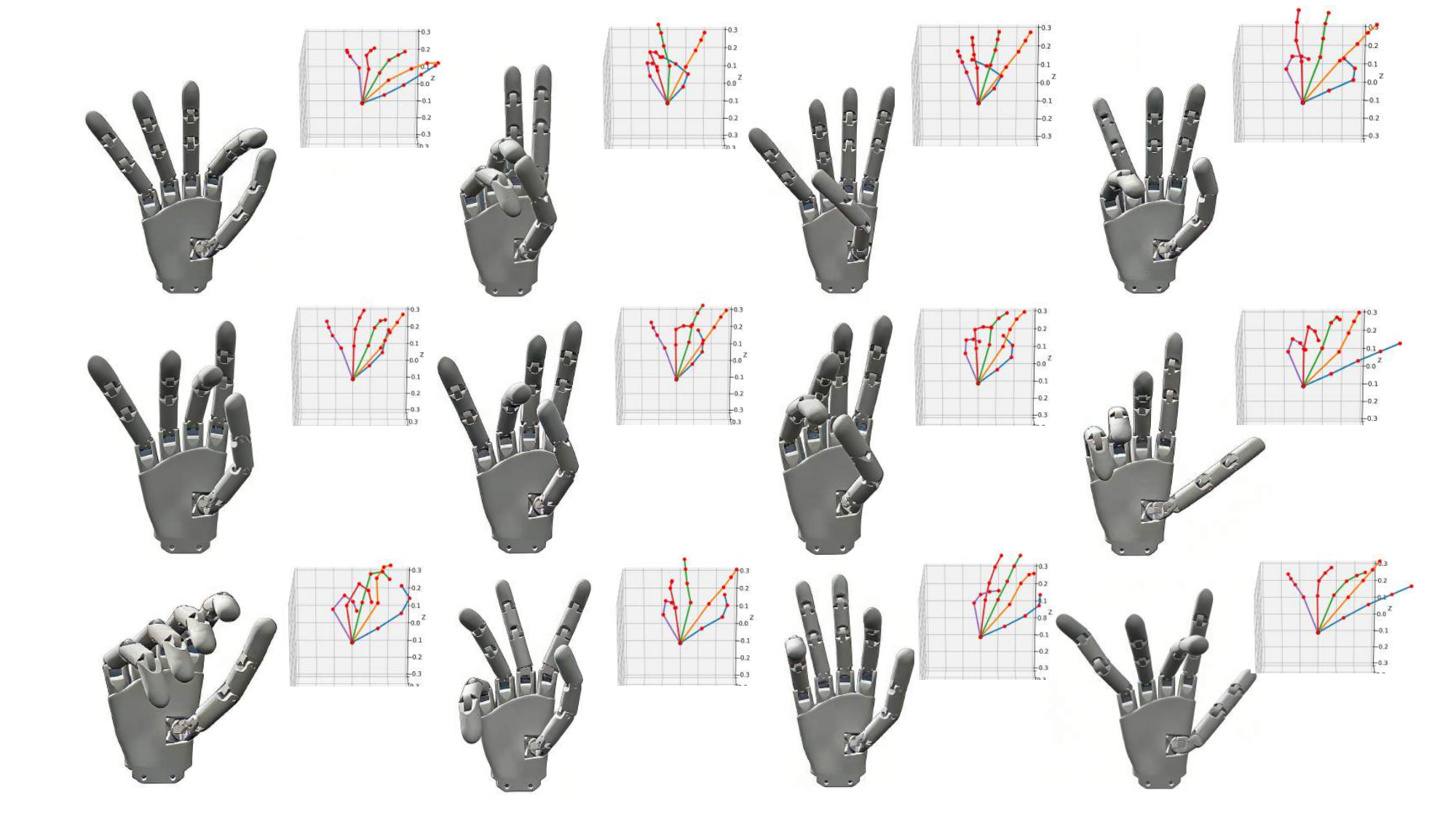}
  \caption{Qualitative results illustrating robust human-to-robot hand retargeting across diverse hand gestures.}
  \label{fig3}
\end{figure}

Furthermore, we conduct systematic experimental evaluations under a diverse set of human hand poses to assess the robustness of the proposed method across varying pose configurations. Fig. \ref{fig3} illustrates several representative retargeting results. As can be observed, our method is able to stably retarget a wide range of human hand poses, demonstrating strong generalization capability.

\begin{figure}[!htb]
  \centering
  \hspace*{-0.9cm} 
  \includegraphics[width=0.55\textwidth]{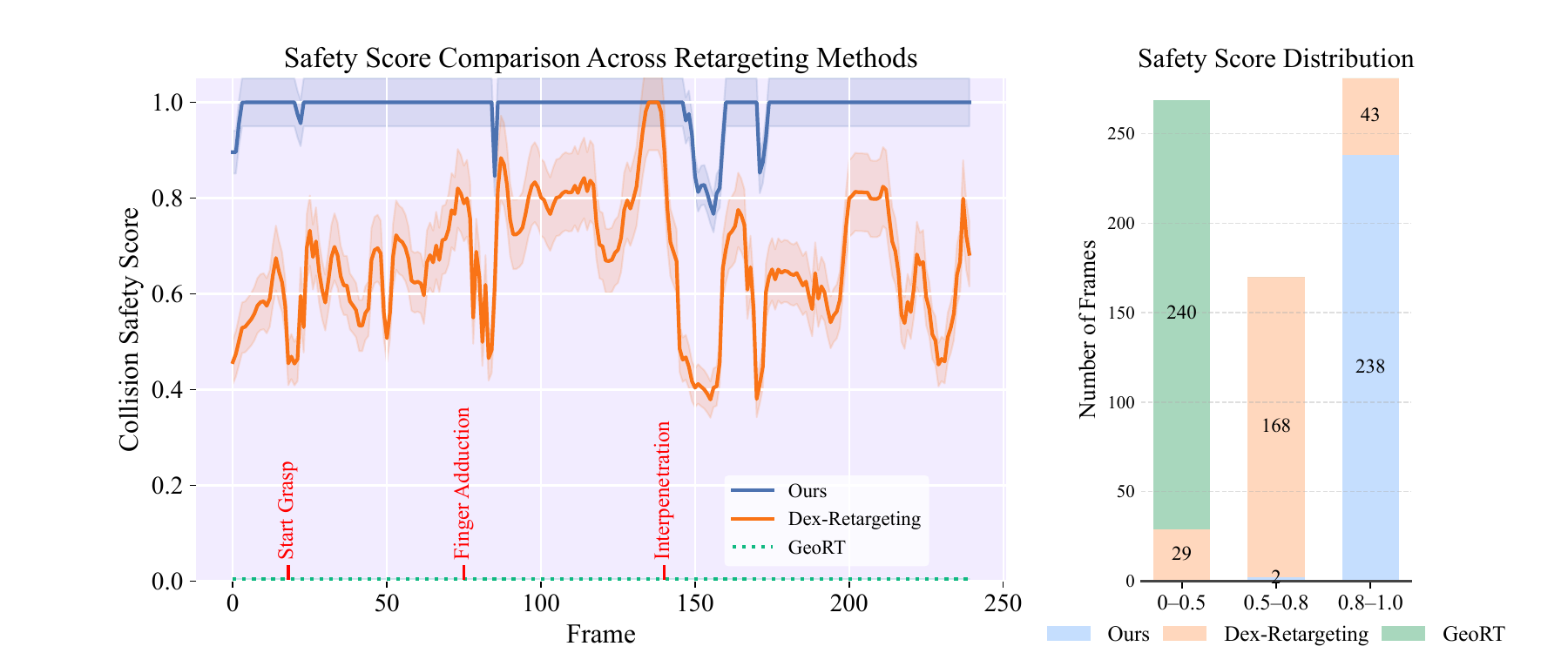}
  \caption{\textbf{Analysis of Safety Score Across Different Retargeting Pipelines.} Annotated stages are defined based on the human hand motion and are shared across all methods.}
  \label{fig4}
\end{figure}
\subsubsection{Safety Analysis}
Collision safety is critical for dexterous hand retargeting, particularly during dense inter-finger interactions. Fig.~\ref{fig4} plots the safety score over time with key stages annotated; all methods are temporally aligned for a fair comparison. Our approach consistently maintains higher safety scores, whereas Dex-Retargeting and GeoRT show sharp degradations during finger adduction and other interpenetration-prone phases, due to the absence of explicit collision-aware constraints. Quantitatively, more than 95\% of control steps achieve a safety score above 0.8, indicating sustained collision clearance without sacrificing real-time performance or motion fidelity.

\begin{figure}[!htb]
  \centering
  \includegraphics[width=0.5\textwidth]{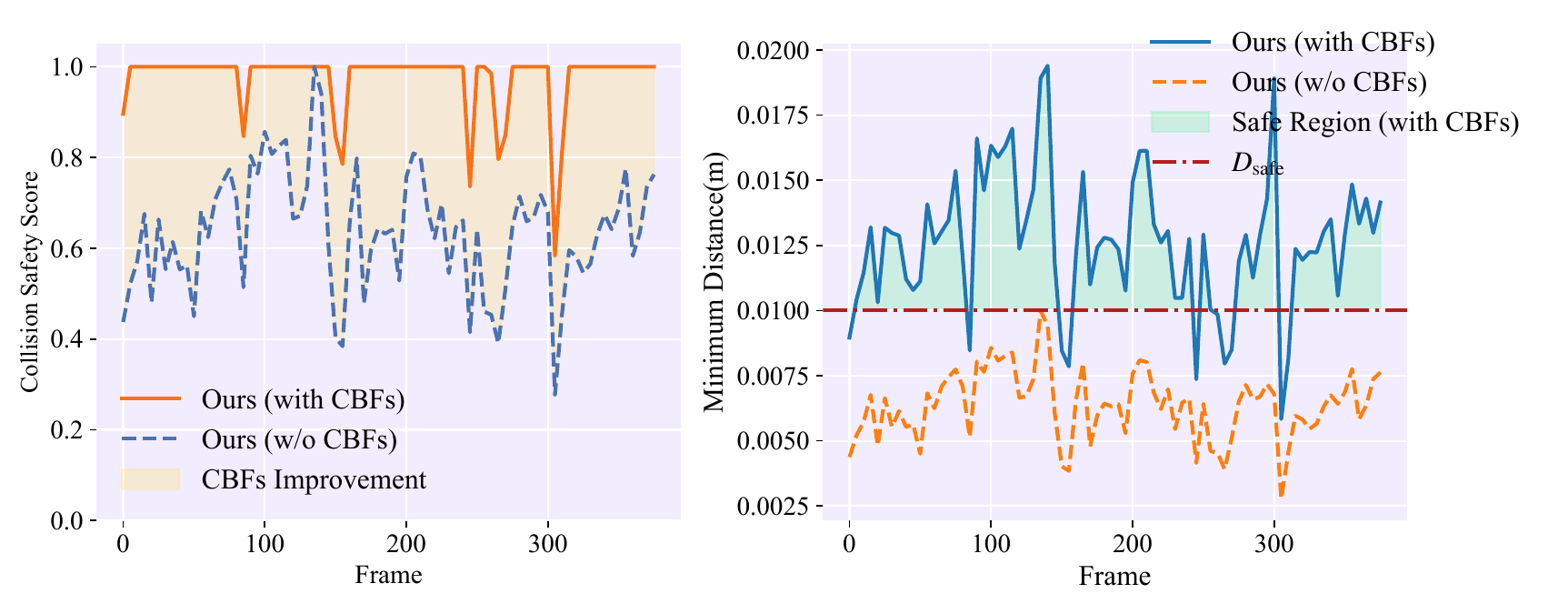}
  \caption{\textbf{CBF ablation study comparing the safety score and minimum inter-finger distance with and without CBF constraints. }The CBF-constrained method maintains consistently higher safety levels, with only minor millimeter-scale threshold violations in a few frames due to discrete-time control and system latency.}
  \label{fig5}
\end{figure}

To isolate the contribution of the explicit safety terms, we perform an ablation that removes the safety terms while keeping all other components unchanged. Under frame-aligned tracking (the retargeted motion follows the human input at each time step), Fig.~\ref{fig5}(a) shows that the safety score frequently drops below the prescribed threshold without safety constraints, indicating elevated collision risk. Enforcing the safety constraints, in contrast, yields consistently high scores throughout the sequence.

We further analyze the induced physical safety behavior by comparing the per-frame minimum inter-finger distance (the minimum distance over all finger pairs) with and without safety constraints (Fig.~\ref{fig5}(b)). With safety constraints enabled, the decrease in inter-finger distance is actively attenuated as it approaches the safety threshold, effectively preventing imminent collisions.

We set the minimum safety distance threshold to 0.01~m, which defines the normalized safety score. The threshold is selected according to the Wuji hand’s finger thickness and joint clearances and is kept fixed across all experiments. To account for system latency, we additionally introduce an activation distance that triggers the safety constraints in advance. A larger activation distance yields earlier intervention and thus more proactive collision avoidance. In our experiments, the activation distance is set to 0.011~m, only slightly above the safety threshold. During grasping, inter-finger distances can decrease rapidly; once the distance falls below the activation threshold, the constraints become active and explicitly restrict finger motions to avoid collisions and penetrations. Notably, even with this minimal anticipation margin, the proposed formulation effectively suppresses further distance reduction under latency.

\begin{figure}[H]
  \centering
  \hspace*{-0.5cm} 
  \includegraphics[width=0.53\textwidth]{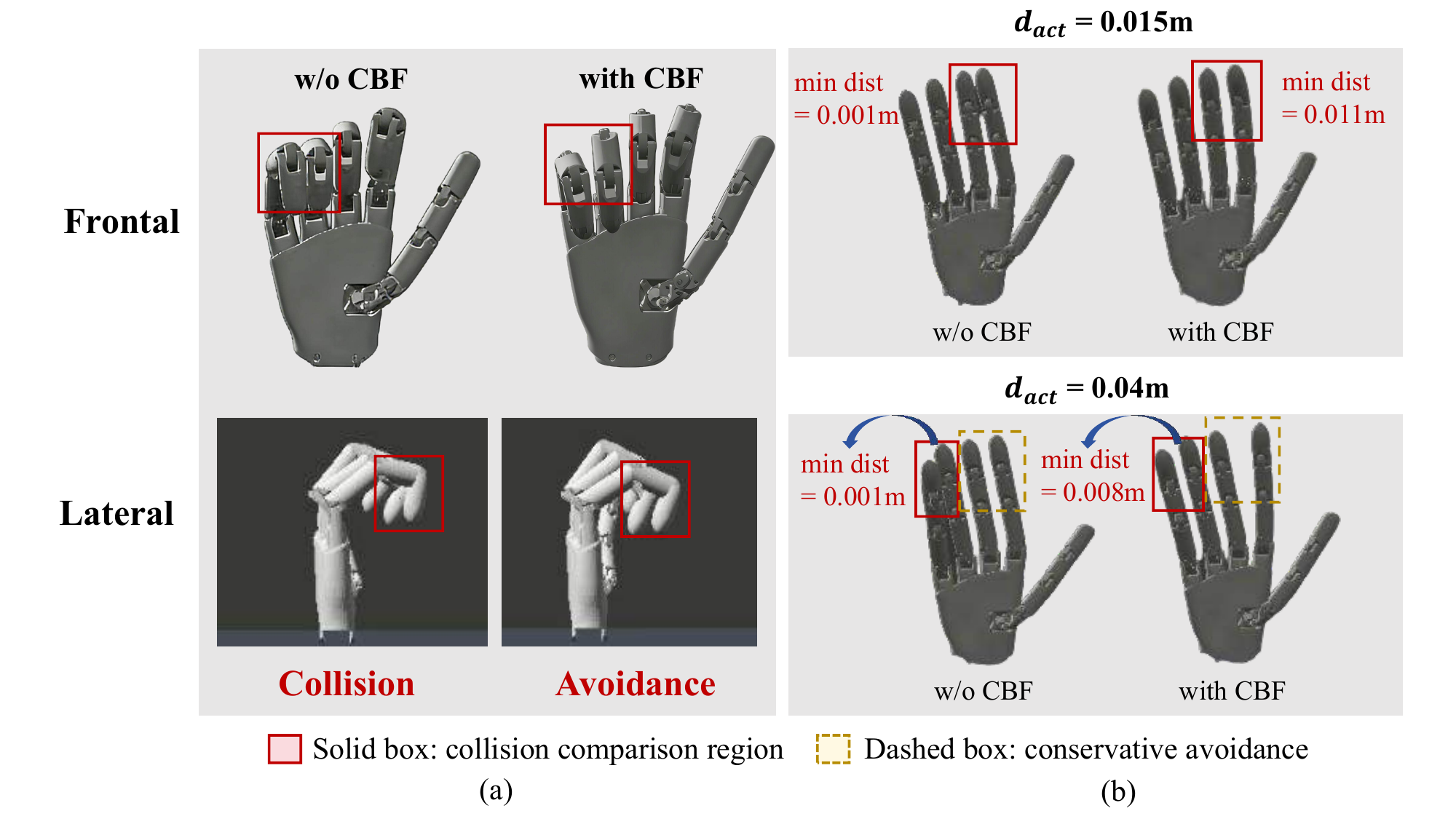}
  \caption{\textbf{Comparison of safety evaluation results for hand grasping and inter-finger contact motions with and without safety constraints.}
(Left) Front and side views of a grasping motion, illustrating how safety constraints prevent inter-finger collisions and geometric interpenetration.
(Right) Effects of increasing activation distance, where larger activation distances lead to earlier intervention and more pronounced collision avoidance.}
  \label{fig6}
\end{figure}

Finally, we qualitatively evaluate safety-constrained retargeting on representative gestures (Fig.~\ref{fig6}). During grasping (left), when the human fingers move into close or overlapping configurations, safety constraints limit unsafe relative motions between the corresponding robot fingers and prevent geometric interpenetration; removing the constraints leads to frequent self-collisions and severe penetration artifacts. Similar issues occur during inter-finger contact (right): without explicit safety constraints, interpenetration becomes pronounced, while enforcing the constraints adaptively adjusts joint motions to satisfy safety requirements while preserving the overall gesture characteristics. Increasing the activation distance further strengthens the preventive effect by intervening prior to contact. 

\subsection{Real-World Evaluation}
In hardware experiments, we focus on qualitative evaluation to assess real-world feasibility and motion naturalness. A series of real-time human hand motions are tested, including grasping and finger-crossing behaviors that are common in daily manipulation and impose strict requirements on responsiveness and collision avoidance. During evaluation, we primarily examine the real-time responsiveness and safety performance of the dexterous hand while executing continuous motion sequences, validating the proposed framework under practical sensing and actuation conditions.

\begin{figure}[!htb]
  \centering
  \includegraphics[width=0.5\textwidth]{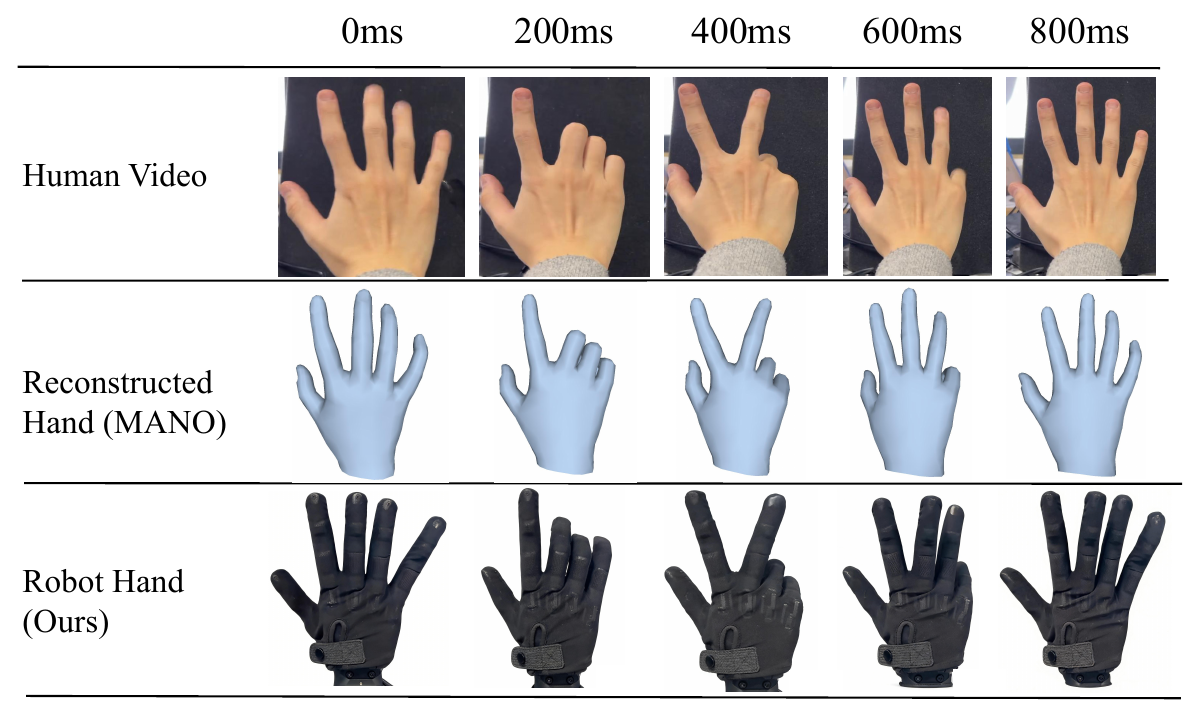}
  \caption{\textbf{Temporal snapshots sampled every 200ms demonstrating real-time motion retargeting.}}
  \label{fig7}
  \end{figure}

The proposed framework maintains fixed-rate closed-loop execution during hardware deployment, enabling consistent control updates and continuous motion retargeting. The robot hand smoothly follows human motions over extended continuous sequences without observable control interruptions or delay accumulation, indicating stable and reliable real-time operation under practical sensing and actuation conditions, as shown in  Fig. \ref{fig7}.

\begin{figure}[!htb]
  \centering
  \includegraphics[width=0.5\textwidth]{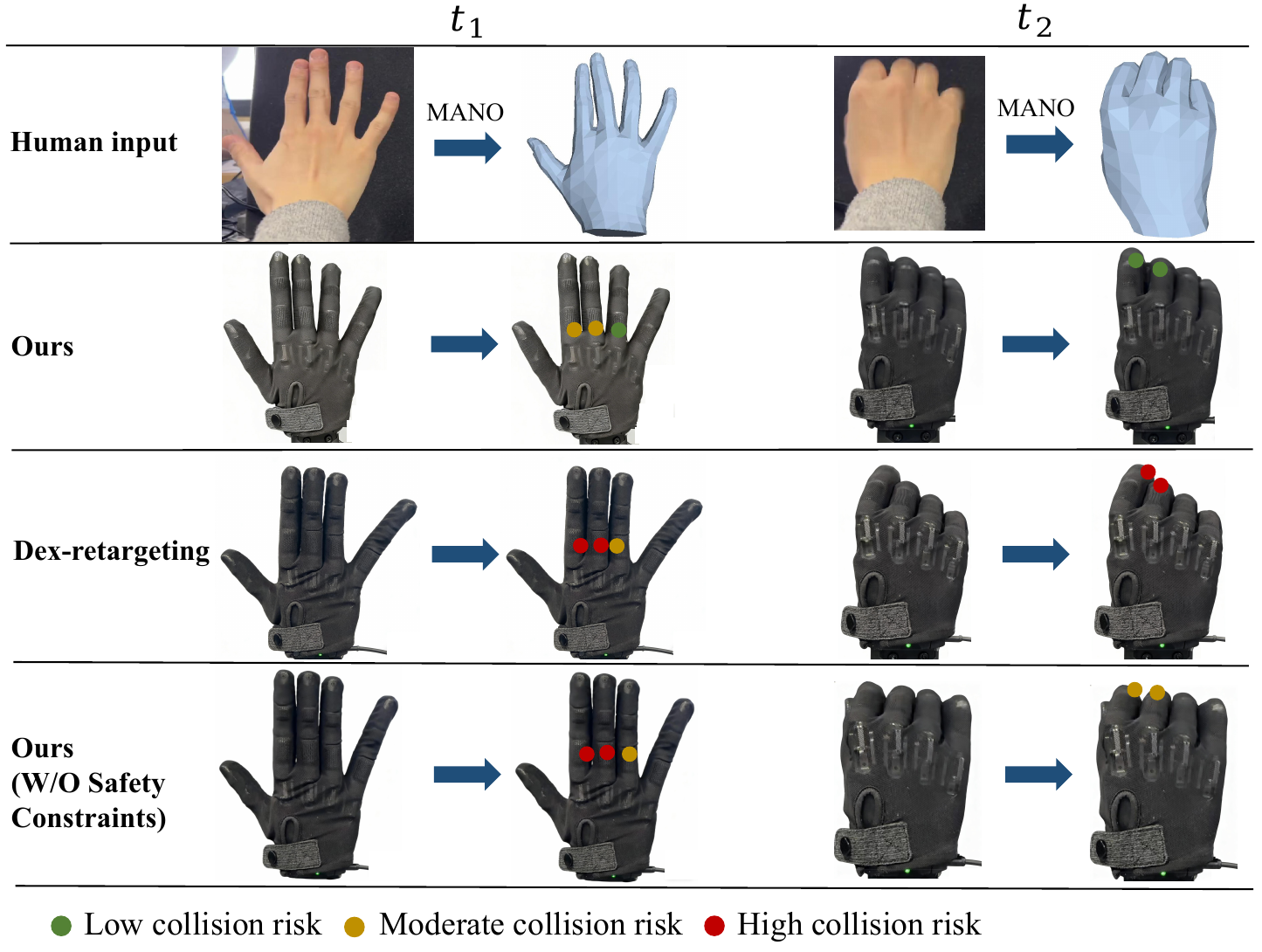}
  \caption{\textbf{Qualitative comparison of collision risk in the inter-finger region during hand closing.} Colored markers indicate the collision risk level at specific inter-finger locations (green: low, yellow: moderate, red: high). Our method maintains lower collision risk compared to Dex-Retargeting and the variant without safety constraints.}
  \label{fig8}
   \end{figure}

To further evaluate execution safety during real-time operation, we qualitatively compare the proposed framework with Dex-Retargeting, and an ablation variant without the QP-based safety constraints under identical motion inputs. As shown in Fig. \ref{fig8}, the baseline method may produce configurations with reduced inter-finger clearance or unstable finger interactions during challenging motions such as finger crossing. In contrast, the proposed framework maintains safe hand configurations throughout execution, with no observable self-collisions while preserving motion continuity and stability. These observations indicate that the introduced safety constraints effectively enable collision-aware motion retargeting during real-world execution.

Overall, the hardware experiments validate the practical deployability of the proposed framework in real-world environments. The presented motions cover diverse articulation patterns including finger coordination, rapid posture transitions, and large configuration changes, providing representative evaluation scenarios for retargeting robustness. The proposed method achieves stable fixed-rate closed-loop execution while maintaining safe hand configurations during challenging motions. Together with the quantitative simulation results, these findings indicate that the proposed approach enables reliable, real-time, and safety-aware motion retargeting under practical sensing and actuation conditions.

\section{CONCLUSIONS}

This paper proposes a dexterous hand mapping framework based on Convex Quadratic Programming, which enables the efficient integration of heterogeneous constraints and significantly enhances computational efficiency. By introducing Control Barrier Functions, the framework provides theoretical safety guarantees, effectively eliminating self-collision risks while preserving motion similarity. Future research will explore the integration of tactile impedance control constraints into the existing framework to optimize torque distribution during object interaction and address the issue of motor torque saturation. 

\addtolength{\textheight}{-12cm}   








\bibliographystyle{IEEEtran}
\bibliography{ref}

\end{document}